\documentclass[10pt,twocolumn,letterpaper]{article}

\usepackage{wacv}
\usepackage{times}
\usepackage{epsfig}
\usepackage{graphicx}
\usepackage{amsmath}
\usepackage{amssymb}
\usepackage{color}
\usepackage{booktabs} % For formal tables
\usepackage{multirow}
\usepackage{array}
\usepackage{verbatim}
\usepackage{caption}
\usepackage{array, siunitx, boldline, hhline}
\usepackage{threeparttable}  

\usepackage{hyperref}
\hypersetup{
    colorlinks=true,
    linkcolor=red,
    filecolor=magenta,      
    urlcolor=red,
}
 
\wacvfinalcopy % *** Uncomment this line for the final submission

\def\wacvPaperID{****} % *** Enter the wacv Paper ID here

\ifwacvfinal\fi
\begin{document}

%%%%%%%%% TITLE

\title{Ancient Painting to Natural Image: A New Solution for Painting Processing}

\author{Tingting Qiao\qquad Weijing Zhang\qquad Miao Zhang\qquad Zixuan Ma\qquad Duanqing Xu\\ Zhejiang University, China \\ \tt\small {\{qiaott, zhangweijing, zhangmiao, aakk883, xdq\}@zju.edu.cn}}
%\author{
%Tingting Qiao \and Jianfeng Dong \and Duanqing Xu \\
%Zhejiang University, China \\ 
%\{qiaott, danieljf24, xdq\}@zju.edu.cn

%
%\author{Tingting Qiao \\
%\and
%Weijing Zhang \\
%\and
%Miao Zhang \\
%\and
%Zixuan Ma \\
%\and
%Duanqing Xu \\
%}

\maketitle
\ifwacvfinal\thispagestyle{empty}\fi

\begin{abstract}
Collecting a large-scale and well-annotated dataset for image processing has become a common practice in computer vision. 
However, in the ancient painting area, this task is not practical as the number of paintings is limited and their style is greatly diverse. 
We, therefore, propose a novel solution for the problems that come with ancient painting processing. 
This is to use domain transfer to convert ancient paintings to photo-realistic natural images. 
By doing so, the \emph{"ancient painting processing problems"} become \emph{"natural image processing problems"} and models trained on natural images can be directly applied to the transferred paintings. 
Specifically, we focus on Chinese ancient flower, bird and landscape paintings in this work. 
A novel Domain Style Transfer Network (DSTN) is proposed to transfer ancient paintings to natural images which employ a compound loss to ensure that the transferred paintings still maintain the color composition and content of the input paintings. 
The experiment results show that the transferred paintings generated by the DSTN have a better performance in both the human perceptual test and other image processing tasks than other state-of-art methods, 
indicating the authenticity of the transferred paintings and the superiority of the proposed method. 
\end{abstract}

\section{Introduction}
\begin{figure}[tb!]
\centering
\noindent\includegraphics[width=0.9\columnwidth]{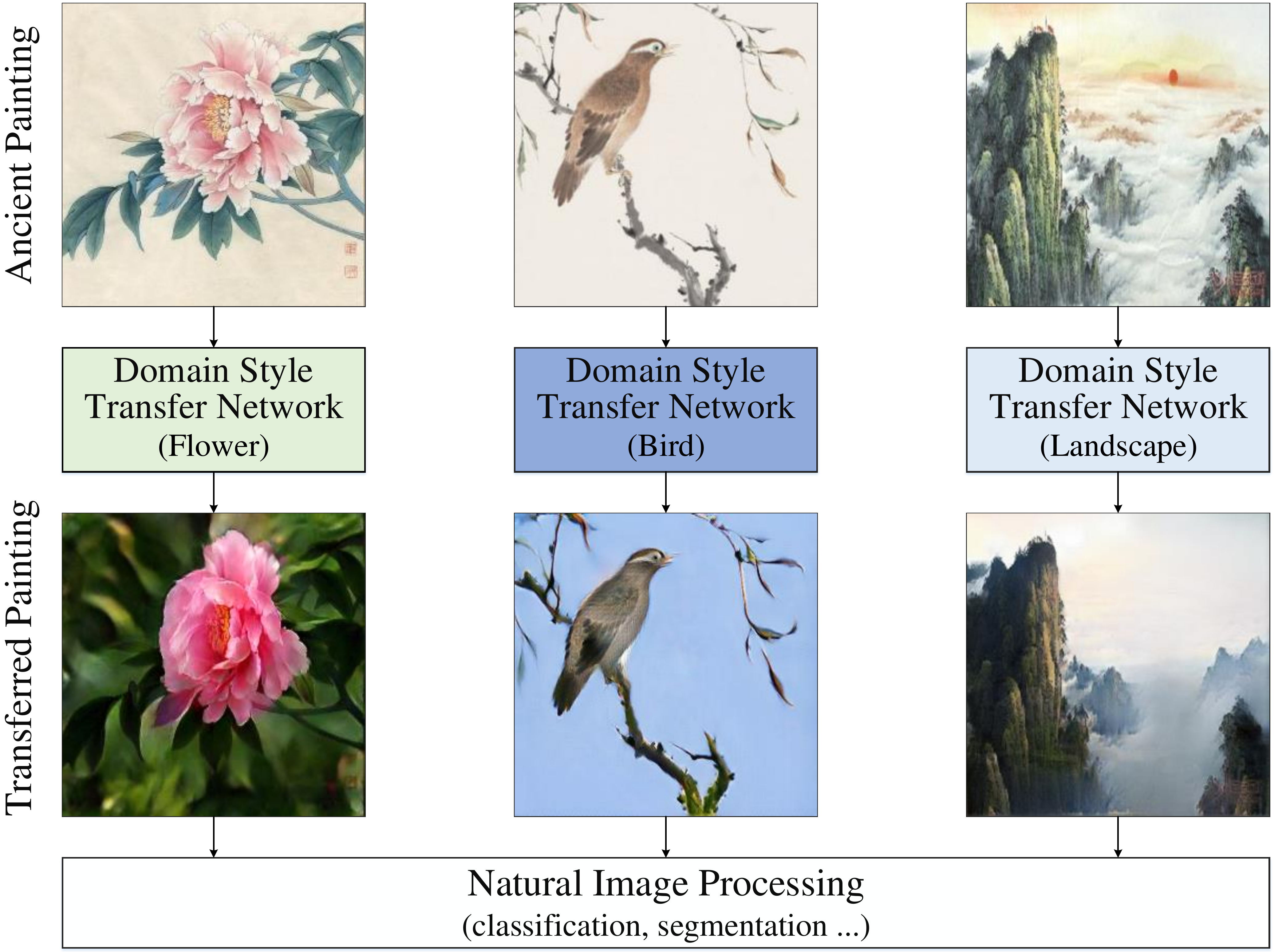}
\protect\caption{General idea. We present a novel solution for ancient painting processing. 
Ancient paintings are first transferred to photo-realistic natural images by the proposed Domain Style Transfer Network (DSTN). 
Then, the models trained on the natural image domain can be directly applied to transferred paintings. By doing so, painting processing problems become natural image processing problems. }
\label{fig:idea}
\end{figure}

Image processing is a fundamental topic in the field of computer vision and pattern recognition. 
Published large and well-annotated datasets, 
such as ImageNet \cite{deng2009imagenet} and COCO dataset \cite{lin2014microsoft}, 
which contain a large number of labeled images, 
have encouraged the development of image processing research, 
such as image classification \cite{krizhevsky2012imagenet,szegedy2015going,simonyan2014very,he2016deep} and semantic segmentation \cite{long2015fully,chen2017rethinking,hariharan2011semantic}. 
However, in the field of ancient paintings, collecting a large scale and well-annotated dataset is not practical. 
This is due to the fact that the number of paintings is limited, and also because ancient paintings from different eras and painters differ greatly in style, color and quality. 
Therefore, painting processing has always relied on manual work which habitually consumes a lot of time. 
One possible way to tackle painting processing problems is to directly apply models trained on natural images to ancient paintings. 
However, this leads to bad performance due to the domain bias. 
This paper demonstrates a novel solution for ancient painting processing, 
which is to use the domain transfer technique to convert ancient paintings to natural images. 
In this way, the rich knowledge learned from the natural image domain can be applied to the transferred paintings and the \emph{"painting processing problems"} can be tackled as \emph{"natural image processing problems"}, which has been widely studied. 
The illustration of our general idea is shown in Figure \ref{fig:idea}. 

In this work, we mainly focus on Chinese ancient flower, bird and landscape paintings. 
Three Chinese ancient painting datasets are first manually collected, namely Chinese Flower Painting dataset (CFP), Chinese Bird Painting dataset (CBP) and Chinese Landscape Painting dataset (CLP), 
%which contain around 4,000, 2,700 and 2,600 paintings, respectively. 
%Besides this, around 4,500 natural flower images, 3,500 natural bird images and 3,500 natural landscape images are also collected as the corresponding natural image domain. 
Then, based on Generative Adversarial Nets (GANs) \cite{goodfellow2014generative}, a novel Domain Style Transfer Network (DSTN) is proposed, 
which is able to transfer input ancient paintings to plausible natural images, 
while ensuring that the transferred paintings maintain their initial content and color composition. 
Three DSTNs are trained using the unpaired images contained in the collected datasets, 
\emph{$\{$flower painting $\rightarrow$ natural flower image$\}$}, \emph{$\{$bird painting $\rightarrow$ natural bird image$\}$} and \emph{$\{$landscape painting $\rightarrow$ natural landscape image$\}$} respectively. 

To verify the authenticity of the transferred paintings and the practicability of the proposed method, 
a variety of experiments is conducted on transferred paintings, 
including a human perceptual study, fine-grained classification on transferred flower paintings and semantic segmentation on transferred bird paintings. 
%In addition, the proposed network is compared with other recent state-of-art generative models. 
Both qualitative and quantitative results show the authenticity of the transferred paintings and the superiority of the proposed method. 
To summarize, the main contributions of our work are as follows: 

1. We propose a novel solution for ancient painting processing problems which is to use the domain transfer technique to convert ancient paintings to natural images, so that the models trained on natural images can be directly applied to transferred paintings. 

2. We present a novel GANs-based Domain Style Transfer Network (DSTN), which successfully transfers ancient paintings to plausible natural images in an unsupervised manner, 
while ensuring that the transferred paintings still maintain their initial content and color composition. 
Our method is not limited to the domain transfer between ancient paintings and natural images, but could as well be applied to other domain transfer tasks where paired data is not available. 

3. Three Chinese ancient painting datasets (CFP, CBP, CLP) are collected which can further be used for other tasks, 
such as painter identification and style identification. 
The \href{https://github.com/qiaott/AncientPainitng2NaturalImage}{datasets} has been made available to the public. 

\section{Related Work}
\subsection{Ancient Painting Processing}
In the past few years, the digitalization and the publication of fine-art collections have been increasing rapidly. 
There are several works which try to extract information from these collections. 
\cite{yang2017painting} uses an online learning algorithm to classify painting images by using multi-features. 
\cite{chen2017multi} focuses on ancient paintings' chronological classification problems by extracting a uniform feature that can represent the multiview
appearance and color attributes of objects and use this feature for ancient paintings chronological classification. 
\cite{agarwal2015genre} explores the problem of feature extraction on paintings and focuses on the classification of paintings into their genres and styles. 
Different from the mentioned works, we aim to use the domain transfer technique to convert ancient paintings to natural images. 
In this way, the diverse styles of the ancient paintings are removed and the rich knowledge learned from natural images can be directly applied to the transferred ancient paintings, which indirectly advances the development of research on ancient paintings. 

\begin{figure*}[tb!]
\centering
\noindent\includegraphics[width=2.0\columnwidth]{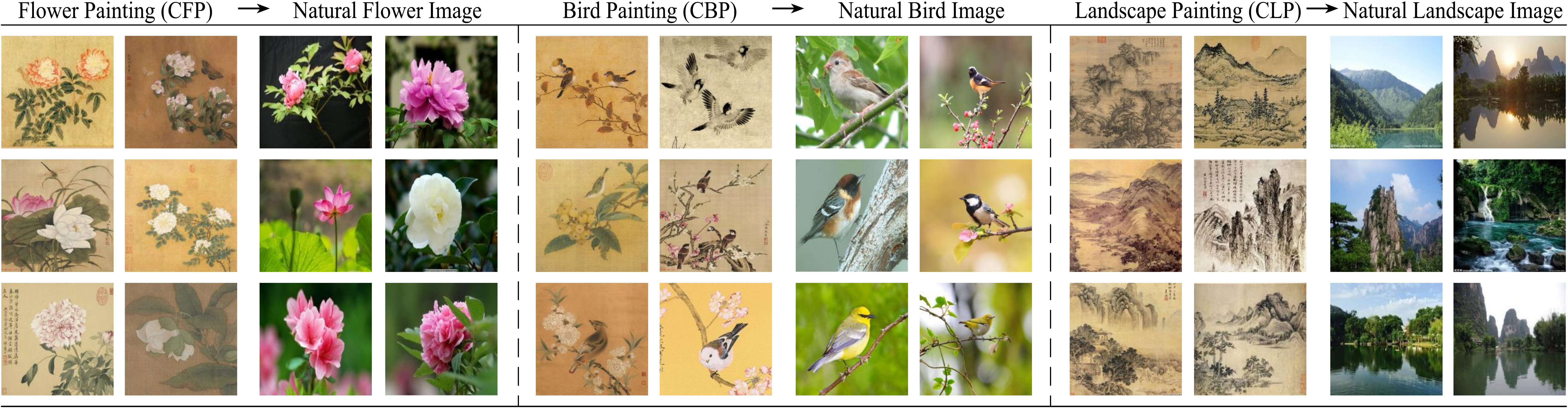}
\protect\caption{Examples of datasets collected for domain style transfer.}
\label{fig:datasets}
\end{figure*}
\subsection{Generative Adversarial Networks}
Generative Adversarial Nets (GANs) \cite{goodfellow2014generative} are a recent popular method for learning generative models of high-dimensional unstructured samples. 
%GANs consist of two networks, 
%a generative model $G$ that generates fake samples resembling real samples and a discriminative model $D$ that distinguishes between samples generated by $G$ and samples drawn from the training data by predicting a binary label. 
%Competition in this process drives both networks to improve their methods, 
%ultimately resulting in the generator $G$ converging to the data distribution of the real samples. 
GANs have performed very well in a wide variety of applications such as image generation \cite{denton2015deep}, super-resolution \cite{ledig2016photo}, domain transfer \cite{taigman2016unsupervised,bousmalis2016unsupervised,kim2017learning,benaim2017one} and 
image translation \cite{isola2016image,zhu2017unpaired,yi2017dualgan}. 
A representative application is the conditional GANs \cite{mirza2014conditional}, 
which are trained using paired training data and produce satisfactory results. 
However, as paired data is not always available, 
unsupervised image to image translation is proposed \cite{zhu2017unpaired,yi2017dualgan,kim2017learning}, 
where a cycle consistency loss is commonly used. 
Our work aims to realize the task-specific domain transfer between the ancient painting domain and the natural image domain. 
The biggest concern is to ensure that the transferred paintings still preserve the color composition and content of the input paintings. 
In the proposed DSTN, besides an adversarial loss, a compound loss is employed to ensure that transferred paintings still maintain the color composition and content of the input paintings, resulting in aesthetically better and more detailed images. 

\subsection{Domain Transfer}
Domain transfer is frequently used in order to overcome the limitation of large data requirements typically needed for deep learning. 
Extensive work has been done in the domain transfer area. 
The Domain-Adversarial Neural Network \cite{ajakan2014domain,ganin2015unsupervised,ganin2016domain} proposes to extract domain-invariant features, which is achieved by jointly optimizing the underlying features as well as two discriminative classifiers operating on the label predictor and the domain classifier. 
%The work in \cite{yoo2016pixel} transfers an input domain to a target domain in semantic level and generates the target image in pixel level. 
\cite{tzeng2017adversarial} proposes Adversarial Discriminative Domain Adaptation which combines discriminative modeling, untied weight sharing and a GANs loss. 
The Domain Transfer Network \cite{taigman2016unsupervised} learns a generative function that maps an input sample from a source domain to a target domain, such that the output of a given function $f$, which accepts inputs in either domain, would remain unchanged. 
On the contrary, we focus on the opposite transfer direction, 
which is to transfer unlabeled paintings in the source domain to natural images in the labeled target domain, 
so that the rich knowledge learned from the natural images can be applied to the transferred paintings. 
The reason why we do not adopt the domain transfer from the natural image domain to the ancient painting domain is the fact that the number of ancient paintings is limited and their styles are greatly diverse. Therefore, in that case, the learned mapping function may have a high-uncertainty. 

\section{Domain Style Transfer} \label{sec:network}
This work presents a novel solution for ancient painting processing problems. 
The ancient painting and natural image datasets created for this task are introduced in Section \ref{sec:dataset}. 
The proposed Domain Style Transfer Network (DSTN), which is used to transfer ancient paintings to photo-realistic natural images, is introduced in Section \ref{sec:DSTN}. 

\begin{table}
\centering
\fontsize{7.5}{10}\selectfont 
\caption{\label{tab:dataset_table}Size of datasets collected for domain style transfer.}
\begin{tabular}{>{\bfseries}c*{4}{c}}\toprule
\multirow{2}{*}{\bfseries Domain} & \multicolumn{2}{c}{\bfseries Training Set} 
                                                           & \multicolumn{2}{c} {\bfseries Testing Set} 
                                                             \\\cmidrule(lr){2-3}\cmidrule(lr){4-5}
                       & \textbf{Painting} & \textbf{Natural Img}       & \textbf{Painting}& \textbf{Natural Img} \\ \midrule
    Flower               &2285      &3546                    & 650          & 1000             \\
    Bird                    &2119       &2512                    &600          & 1000              \\ 
    Landscape        &2009     &2458                    &600          &1000                \\ 
    \bottomrule
\end{tabular}
\end{table}

\subsection{Datasets} \label{sec:dataset}
In this work, we mainly focus on Chinese flower, bird and landscape paintings as they are very important parts of Chinese painting history. 
Three Chinese ancient painting datasets are manually collected, 
namely Chinese Flower Painting dataset (CFP), Chinese Bird Painting dataset (CBP)  and Chinese Landscape Painting dataset (CLP). 
Moreover, corresponding natural images are also collected. 
Samples of the collected three painting datasets and natural images are shown in Figure \ref{fig:datasets}. 
For the sake of completeness, the specific quantity of the training set and the testing set in each dataset is illustrated in Table \ref{tab:dataset_table}. 

\textbf{Collection of Paintings}{\;}
The flower and bird paintings in the CFP and CBP datasets are mainly collected from the Song (960-1279) Dynasty. 
%however there are also a few paintings from the Yuan (1271-1368) and Qing (1636-1912) Dynasties. 
The landscape paintings in the CLP dataset are collected from the Qing Dynasty. 
Paintings are collected from museums, picture albums and Google Image Search. 
For a better transfer effect, each painting in these three datasets is manually cropped in order to remove redundant information
and to focus on the main object. %examples of these datasets are shown in Figure \ref{fig:datasets}. 
As these paintings are selected from different painters in different dynasties, a great diversity is achieved in the datasets. 

\textbf{Collection of Natural Images}{\;}
The natural flower images are collected from Google Image Search. 
The natural bird images are selected from the Caltech-UCSD Birds200 dataset \cite{wah2011caltech}, 
which is a large collection of 200 categories of birds. 
All flower and bird images are chosen to be flowers and birds commonly occurring in the ancient paintings, 
$e.g.,$ peony, camellia, sunflower etc. 
The landscape images are automatically retrieved from Google Image Search. 
Note that there are no paired images between any of the two domains. 

\subsection{Domain Style Transfer Network}\label{sec:DSTN}
\begin{figure*}[tb!]
\centering
\noindent\includegraphics[width=1.90\columnwidth]{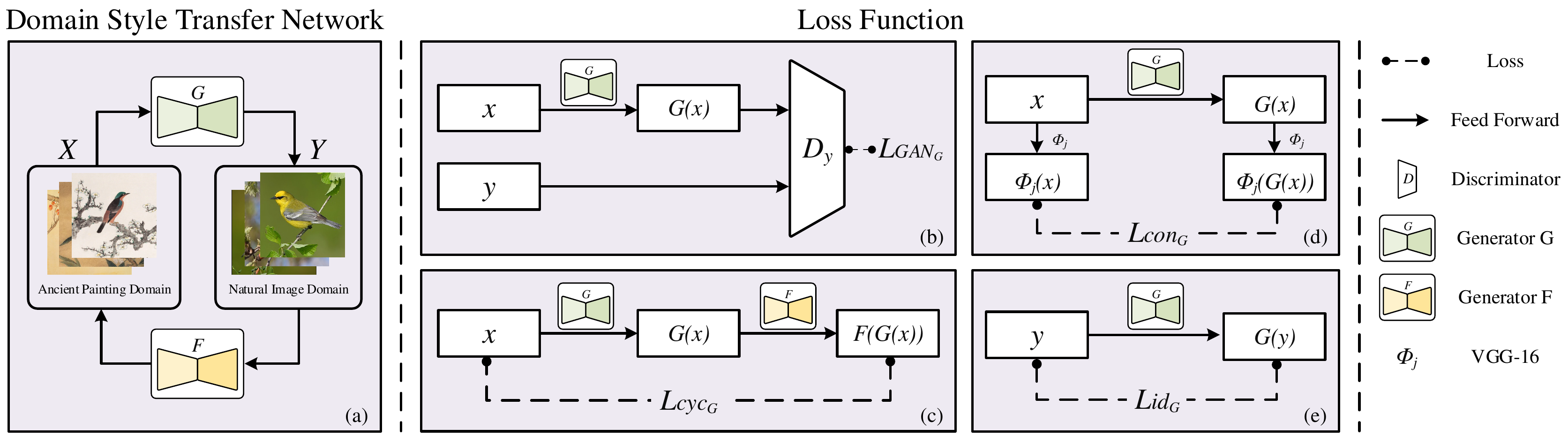}
\protect\caption{(a) Structure of the Domian Style Transfer Network (DSTN), in which two mapping functions $G$ and $F$ between two domains are learned. 
For the generator $G$ and its corresponding discriminator $D_y$, a compound loss, which consists of (b) $L_{GAN_{G}}$, (c) $L_{cyc_{G}}$,  (d) $L_{con_{G}}$ and (e) $L_{id_{G}}$, is employed.}
\label{fig:framework}
\end{figure*}

Given a source domain consisting of ancient paintings and a target domain consisting of natural images,
the goal of this network is to transfer ancient paintings to the natural image domain so that the transferred paintings can be regarded as real natural images. 
As inspired by the structure proposed in \cite{zhu2017unpaired}, the proposed GANs-based Domain Style Transfer Network (DSTN) adopts two pairs of generator-discriminator to ensure the cycle-consistency of the transfer cycle between input and output. 
In addition, it is necessary to ensure the consistency of color composition and to maintain the original painting content during the domain transfer operation. 
Therefore, a compound loss, which combines a cycle consistency loss, a content loss and an identity loss together, is used to make the network capable of generating images that still maintain the original color composition and content of input paintings. 
The details of the network are introduced in the following. 

\textbf{Adversarial Loss}{\;}
The two generator-discriminator pairs employed are represented as $\{G, D_{y}\}$ and $\{F, D_{x}\}$. 
Each generator is trained jointly with its corresponding discriminator, 
which distinguishes between real samples and generated samples. 
More formally, let $X = \{{x_i}\}_{i=1}^{N^x}$ represent $N^{x}$ samples from the ancient painting domain $X$ 
and $Y = \{{y_j}\}_{j=1}^{N^y}$ represent $N^{y}$ samples from the natural image domain $Y$. 
Two mapping functions between both domains are learned during the training, 
which are $G: X \rightarrow Y$ and $F: Y \rightarrow X$. 
For the generator $G$ and its corresponding discriminator $D_y$, 
the adversarial loss is: 
\begin{equation}
\begin{split}
L_{GAN_{G}}=\mathbb{E}_{y\sim p_{data}(y)}[logD_y(y)] \\
+\mathbb{E}_{x\sim p_{data}(x)}[log(1-D_y(G(x))] \label{eq:G}
\end{split}
\end{equation}
where $p_{data}(x)$ and $p_{data}(y)$ represent the data distributions in the $X$ and $Y$ domains respectively. 
The generator $G$ tries to generate images G(x) that look similar to images from the domain Y, while $D_y$ aims to distinguish between generated images $G(x)$ and real samples from the domain $Y$. 
The generator $G$ is trained together with the discriminator $D_y$  by solving: 
${G}^{*}=arg\:min_{G}\:max_{D_y}L_{GAN_{G}}$, 
where the generator $G$ tries to minimize Eq.\ref{eq:G} and the discriminator $D_y$ tries to maximize it. 
A similar adversarial loss is also applied to the generator $F$ and its corresponding discriminator $D_x$. 
The total adversarial loss for both generator-discriminator pairs is: 
\begin{equation}
\begin{split}
L_{GAN}=L_{GAN_{G}} + L_{GAN_{F}}\label{eq:GAN}
\end{split}
\end{equation}

\textbf{Cycle Consistency Loss}{\;}
As in most situations, paired datasets are not available, 
which makes the transfer from domain $X$ to domain $Y$ have multi-mappings. 
A cycle consistency loss introduced in \cite{zhu2017unpaired} is proved effective to recover the input images after an image translation cycle and to avoid the well-known problem of mode collapse \cite{goodfellow2014generative}. 
However, the original CycleGAN fixes the weight of adversarial losses and the two parts of the cycle consistency have the same weight. 
We find that different datasets show a difference in susceptibility to the mode collapse problem. 
In addition, the ratio of cycle loss and adversarial loss should be different for different datasets. 
Therefore we introduce weights for different cycle passes. 
To be specific, for each image $x$ in domain $X$, 
the image translation cycle should satisfy $x\rightarrow G(x)\rightarrow F(G(x))\approx x$, 
which is defined as a forward cycle consistency loss: 
\begin{equation}
\begin{split}
L_{cyc_{G}}=\mathbb{E}_{x\sim p_{data}(x)}[\parallel F(G(x))-x \parallel_{1}]
\end{split}
\end{equation}
where the $\parallel .\parallel_{1}$ is the L1 norm. 
A similar backward cycle consistency loss is also applied to the generator $F$ and $G$:
\begin{equation}
\begin{split}
L_{cyc_{F}}=\mathbb{E}_{y\sim p_{data}(y)}[\parallel G(F(y))-y \parallel_{1}]
\end{split}
\end{equation}
The total cycle consistency loss is represented as: 
\begin{equation}
\begin{split}
L_{cyc}=\alpha_G L_{cyc_{G}}+\alpha_F L_{cyc_{F}}
\end{split}
\end{equation}
where $\alpha_G$ and $\alpha_F$ represent the weights for the forward and backward cycle consistency loss. 

\textbf{Identity Loss}{\;}
The generator $G$ transfers an ancient painting to a natural image, 
which can be regarded as removing the paintings transfering the style of the natural image domain to the paintings. 
Imagine that when the input of the generator $G$ is a natural image, as the input already has the style of the natural image, 
the the output should be itself. 
Based on this assumption, 
we employ an identity loss to regularize the generator $G$ to be an identity mapping matrix on $y$ and the generator $F$ to be an identity mapping matrix on $x$, as illustrated in Figure \ref{fig:framework}. 
\begin{equation}
L_{id_{G}}=\mathbb{E}_{y\sim p_{data}(y)}[\parallel G(y)-y\parallel_{1}] 
\end{equation}
A similar identity loss is also employed for the generator $F$. 
The total identity loss is: 
\begin{equation}
L_{id}=L_{id_{G}}+L_{id_{F}}
\label{eq:id}
\end{equation}

\textbf{Content Loss}{\;}
Furthermore, from an ancient painting to a natural image transfer operation, 
it is expected that the transferred painting still preserves the content of the input painting. 
Therefore, a content loss is employed to encourage the input 
and the output to have similar content as motivated by previous works \cite{gatys2016image,johnson2016perceptual,chen2016fast}, 
in which a pre-trained deep convolutional neural network is used to extract the high-frequency content of an image. 
The content loss employed encourages input paintings and output images to have similar high feature representations computed by a loss network $\phi$. 
For the generator $G$, the loss function is represented as the Squared Euclidean Distance between two feature representations: 
\begin{equation}
L_{con_{G}}=\mathbb{E}_{x\sim p_{data}(x)}[\parallel\phi _j(G(x))-\phi _j(x) \parallel _{2}^{2}] 
\label{eq:content}
\end{equation}
where $\parallel .\parallel_{2}$ is the L2 norm and $\phi_j$ stands for the activations of the $j$-th convolutional layer of the network $\phi$, 
which is a 16-layer VGG network pre-trained on the ImageNet \cite{deng2009imagenet} dataset. 
The total content loss is represented as: 
\begin{equation}
L_{con}=L_{con_{G}} + L_{con_{F}}
\end{equation}

The objective function of the proposed DSTN is defined as: 
\begin{equation}
\begin{split}
L(G,F,D_x,D_y)=L_{GAN}+\alpha_G L_{cyc_G}+\alpha_F L_{cyc_F} \\
+\beta L_{id}+ \gamma L_{con}
\end{split}
\label{eq:objective}
\end{equation} 
where $\alpha_G$, $\alpha_F$, $\beta$ and $\gamma$ represent the weights for each loss.
Both generators $G$ and $F$ are trained together under the constraints of the compound loss function.
The goal of the DSTN is to optimize the following objective function: 
\begin{equation}
{G}^{*},{F}^{*}=arg\:\underset{{G,F}}{min} \: \underset{{D_x,D_y}}{max}L(G,F,D_x,D_y)
\end{equation} 
\section{Experiment Setup}
\subsection{Implement Details}
％Following the architectural guidelines of \cite{zhu2017unpaired}, 
The generative networks are based on the architecture employed in \cite{johnson2016perceptual}, 
which consists of two stride-2 convolutions, two fractionally stridden convolutions with stride $\frac{1}{2}$ and nine residual blocks \cite{He2015Deep}. 
The discriminative networks follow the structure of the PatchGAN \cite{isola2016image}, 
which only penalizes the structure at the level of patches and has fewer parameters than a full-image discriminator. 
%Specifically, this discriminator attempts to classify whether a $L\times L$ patch is real or fake and it is trained in a fully convolutional manner across the image, 
%averaging all responses to provide the ultimate output of the discriminator. 
%$L$ is set to 70 as it shows the best performance \cite{isola2016image}. 
The weights $\alpha_G$, $\alpha_F$, $\beta$, $\gamma$ for each loss are set to 1, 0.5, 10, 1, respectively, as they achieve the best performance. 
we compute the content loss at layer $relu2\_2$ of the VGG-16 network $\phi$.
The implementation uses Pytorch \cite{kingma2014adam} and cuDNN \cite{Chetlur2014cuDNN}. 
For more details and experiment reports about weight-chosen and training details, please refer to supplementary materials. 
\subsection{Datasets}\label{sec:DSTNdataset}
The datasets introduced in Section \ref{sec:dataset} are used for training the DSTN. 
All collected paintings and images are resized to $256\times 256$ pixels in this experiment. 
In total, three groups of domain style transfer experiments are conducted, 
namely \emph{$\{$flower painting $\rightarrow$ natural flower image$\}$}, 
\emph{$\{$bird painting $\rightarrow$ natural bird image$\}$} and 
\emph{$\{$landscape painting $\rightarrow$ natural landscape image$\}$}. 
\begin{comment}
\subsection{Training Details}
The weights $\alpha_G$, $\alpha_F$, $\beta$, $\gamma$ for each loss are set to 1, 0.5, 10, 1, respectively, as they achieve the best performance. 
For more details and experiment reports about weight chosen, please refer to supplementary materials. 
In order to reduce the model collapse of original GANs \cite{goodfellow2014generative}, 
the discriminator is updated following the strategy of \cite{shrivastava2016learning}, 
which uses the history of the generated images instead of the images produced by the latest generative network. 
Moreover, for a more stable gradient, 
the negative log-likelihood objective in the original GANs \cite{goodfellow2014generative} is replaced by a least square loss in the \cite{Mao2016}, 
as in \cite{zhu2017unpaired}. 
Adam solver \cite{kingma2014adam} is used with a batch size of 1. 
The learning rate is set to 0.0002 for the first 100 epochs and linearly decreases to zero over the next 100 epochs. 
The initial weights of the network are uniform on a Gaussian distribution with mean equal to 0 and standard deviation of 0.02. 
For computing the content of an image in Eq. \ref{eq:content}, 
we compute the content loss at layer $relu2_2$ of the VGG-16 network $\phi$.
The implementation uses Pytorch \cite{kingma2014adam} and cuDNN \cite{Chetlur2014cuDNN}. 
\end{comment}

\section{Results and Evaluation}
In this section, both qualitative and quantitative experiments are conducted to verify the authenticity of the transferred paintings and the effectiveness of the proposed method. 
First, a comparison is drawn with examples generated by both the proposed DSTN and different methods in Section \ref{sec:visual}. 
Then, a fine-grained flower classification experiment is conducted on the transferred flower paintings in Section \ref{sec:fine-grained_classification}. 
In Section \ref{sec:segm}, a semantic segmentation experiment is carried out on the transferred bird paintings. 
In addition, a human perceptual test is conducted on the transferred flower, bird and landscape paintings in Section \ref{sec:human_study}. 

Besides, the proposed DSTN is compared with three recent state-of-art generative networks, 
namely GANs \cite{goodfellow2014generative}, DualGAN \cite{yi2017dualgan}, CycleGAN and CycleGAN+$L_{id}$ \cite{zhu2017unpaired}. 
The results of GANs are generated using the proposed network by setting $\alpha_G$=$\alpha_F$=$\beta$=$\gamma$= 0.0 in Eq \ref{eq:objective}. 
Note that the GANs used in this work is different from the original GANs \cite{goodfellow2014generative} as it employs a conditional generator. 
The DualGAN and the CycleGAN, which are proposed for the unsupervised image to image translation task, use an adversarial loss and a similar cycle consistency loss but employ different network structures,  
All methods are trained and tested under the same condition \ref{sec:DSTNdataset}. 
Details about each experiment are introduced in corresponding section. 
\subsection{Visualization} \label{sec:visual}
\begin{comment}
\begin{figure}[tb!]
\centering
\noindent\includegraphics[width=1.0\columnwidth]{fig/example}
\protect\caption{Visualization of experiment results generated by the DSTN which successfully transfers ancient paintings to convincing natural images. 
It can be observed that the original content and color composition of the input paintings are well maintained during the transfer. Best viewed in color.}
\label{fig:example}
\end{figure}
\end{comment}

\begin{figure*}[tb!]
\centering
\noindent\includegraphics[width=2.0\columnwidth]{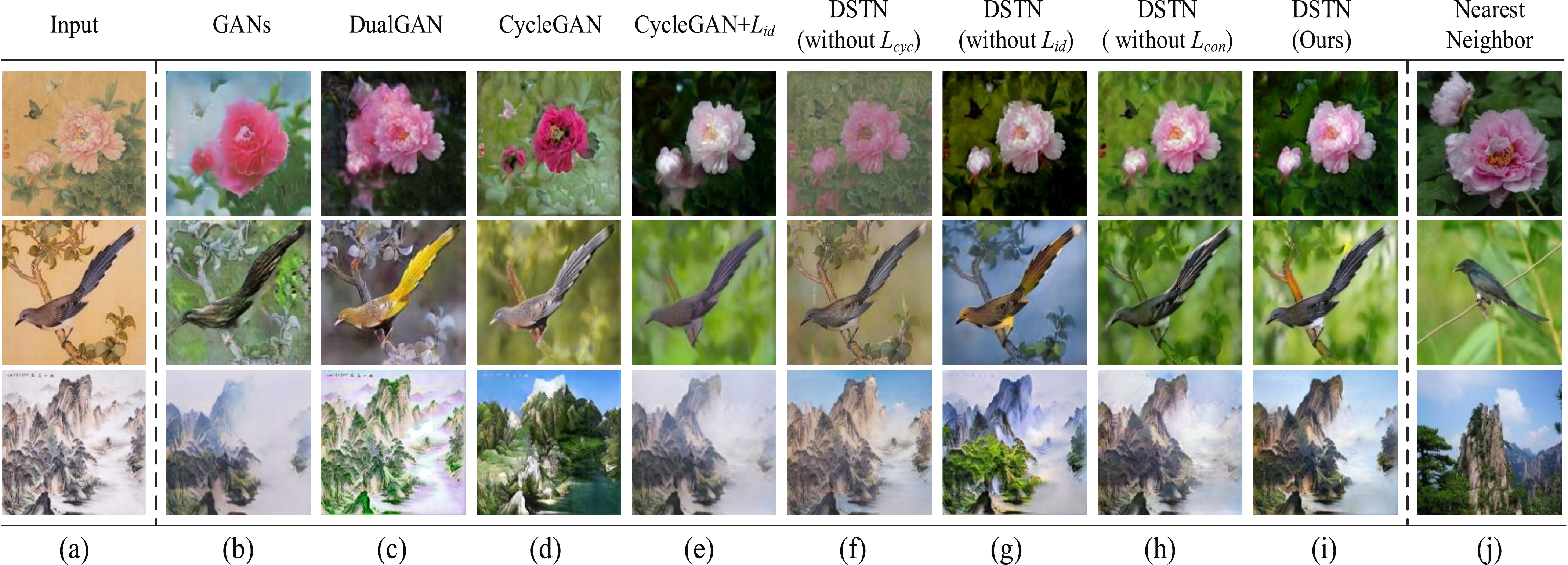}
\protect\caption{Comparison of transferred images produced by different methods and Nearest Neighbor of our generated images. 
The Nearest Neighbors in Column (i) are searched from the target natural image domain, indicating that the DSTN is indeed learning a mapping function instead of merely memorizing the images in the natural image domain. More examples can be found in the supplementary materials.}
\label{fig:compare}
\end{figure*}

\begin{figure*}[tb!]
\centering
\noindent\includegraphics[width=2.0\columnwidth]{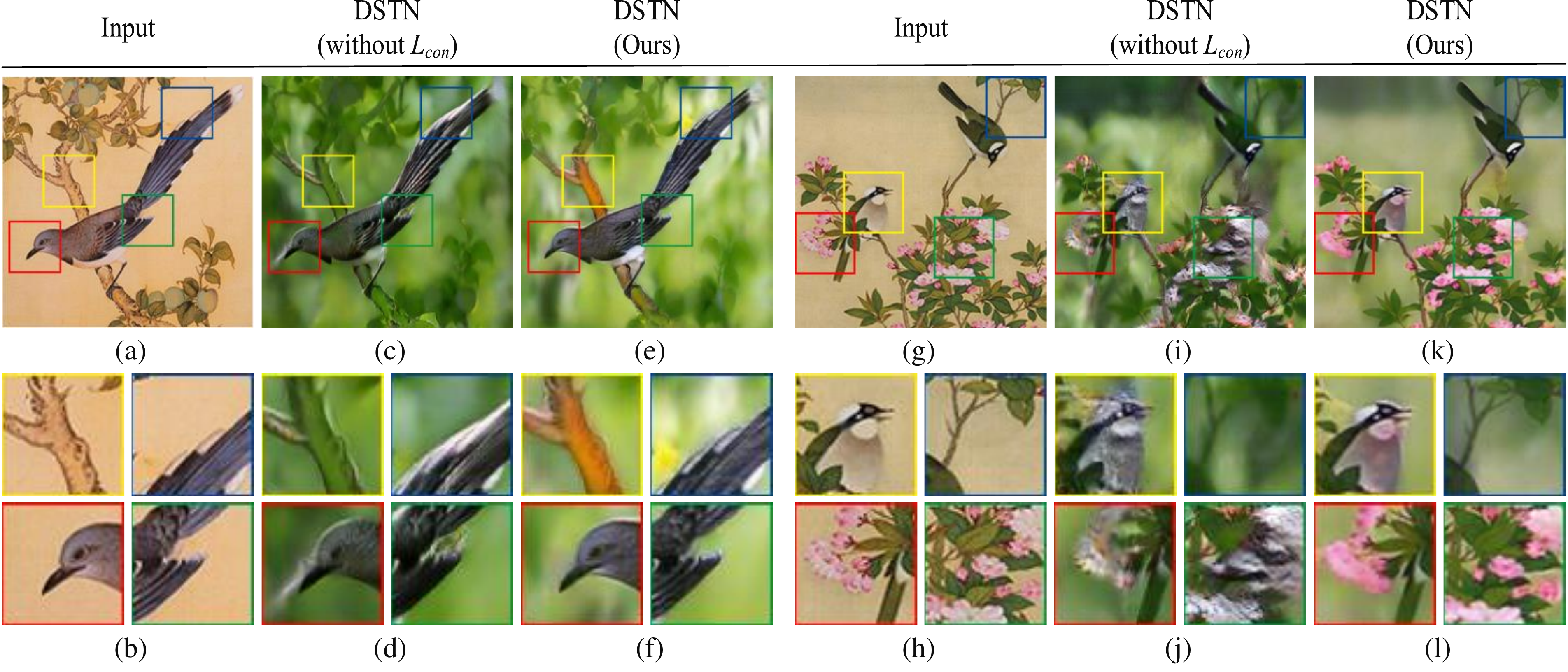}
\protect\caption{Comparison of details of images generated by the DSTN with and without $L_{con}$. Although the generated images produced by both methods look similar in general, the images generated by the DSTN with $L_{con}$ look aesthetically better and more detailed.}
\label{fig:detail}
\end{figure*}

The proposed DSTN is compared with other methods as illustrated in Figure \ref{fig:compare}. 
It can be observed that the images generated by the first three methods, as shown in columns (b, c, d), are relatively coarse and the colors of the generated images are not consistent with the input paintings. 
The DualGAN \cite{yi2017dualgan} and CycleGAN \cite{zhu2017unpaired}, as shown in columns (c, d), are able to generate relatively clear images but with unpredictable colors. 
From the comparison between columns (e, f, g, h), 
it can be observed that each loss term employed in the DSTN contributes to a different effect, resulting in an improvement of the quality of the transferred paintings. 
$L_{id}$ helps to preserve the initial color composition of the input paintings as seen from the comparison between columns (f) and (h), (d) and (e). 
$L_{con}$ helps to generate more authentic and clear images, 
$e.g.,$ in column (g), the main objects of the images generated by the DSTN without $L_{con}$ are not distinctly separate from its surroundings. 
Figure \ref{fig:detail} further shows a comparison between the details of the images generated by the DSTN with and without $L_{con}$. 
It can be observed that the generated images look similar, 
however, the DSTN with $L_{con}$ shows a better performance in dealing with details. 
Overall, the proposed DSTN, which is constrained by a compound loss, is very effective in generating authentic and clear images. 

Furthermore, in order to verify that the model is actually learning a mapping function instead of simply memorizing the images in the target domain, 
we conduct a Nearest Neighbor Lookup in the target domain using the Cosine distance based on the ResNet-152 visual feature \cite{He2015Deep}. 
The Nearest Neighbors of our generated images are illustrated in column (i) of Figure \ref{fig:compare}, showing that the DSTN is not merely memorizing images from the target domain. 

\subsection{Fine-grained Flower Classification} \label{sec:fine-grained_classification}
In this section, a classifier trained on a natural flower dataset is used to classify the transferred flower paintings produced by both the DSTN and other methods. 
which aims to verify the authenticity of the transferred flower paintings and the effectiveness of the proposed method. 
Note that the 650 transferred flower paintings in the testing set of CFP dataset are annotated into 8 categories. 

\textbf{Setup}{\;}
An ImageNet-pretrained \cite{deng2009imagenet} ResNet-152 \cite{He2015Deep} is fine-tuned as a fine-grained natural flower classifier. 
The earlier layers of the ResNet-152 are fixed to avoid overfitting and only the last fully connected layer is fine-tuned. 

\textbf{Dataset}{\;}
The dataset for fine-tuning the fine-grained flower classifier has 9 categories, 
$i.e.,$ 8 categories of flowers needed and others that consist of other categories of flowers, 
the size of each category is shown in Table \ref{tab:category}. 
The flower images are mostly collected from a public dataset Oxford Flower 102 \cite{nilsback2008automated}. 
%Additionally, we collect other types of flower, which are not contained in the Oxford Flower 102, $e.g.,$ peony, malus, peach blossom, from Google Image Search. 
The size of the validation set and the testing set are both set to 2,000 and the rest is used for fine-tuning the classifier. 
%The fine-tuned classifier has a 90.50$\%$ accuracy rate on the testing set for fine-grained natural flower classification. 
%\begin{table}[htbp]
%%\centering
%%\fontsize{7.5}{10}\selectfont
%%\caption{\label{tab:category} Categories of flowers and number of each category in the fine-grained flower classification experiment.}
%%\begin{tabular}{lr|lr}
%%\toprule
%%\textbf{Category} & \textbf{Size}& \textbf{Category} & \textbf{Size}\\
%%\cmidrule{1-4}
%%&lotus                  & {682}                      &magnolia                 & {946}  \\ 
%%&camellia                 & {928}                  & sunflower          & {708}  \\
%%&chrysanthemum    & {947}                  & wintersweet       & {696} \\
%%&peach blossom      & {752}                  & others                  & {1000}\\
%%&peony                 & {559}                      & \textbf{TOTAL}   & {7218} \\
%%\bottomrule
%\end{tabular}
%\end{table}

\begin{table}[htbp]
\centering
\fontsize{7.5}{10}\selectfont
\caption{\label{tab:category} Categories of flowers and number of each category in the fine-grained flower classification experiment.}
\begin{tabular}{lr|lr}
\toprule
\textbf{Category} & \textbf{Size}& \textbf{Category} & \textbf{Size}\\
 \midrule
\cmidrule{1-4}
lotus                  & {682}                      &magnolia                 & {946}            \\ 
camellia                 & {928}                  & sunflower          & {708}  \\
chrysanthemum    & {947}                  & wintersweet       & {696} \\
peach blossom      & {752}                  & others                  & {1000}\\
peony                 & {559}                      & \textbf{TOTAL}   & {7218} \\
\bottomrule
\end{tabular}
\end{table}

\textbf{Experiments}{\;}
The trained models of the DSTN, GANs, DualGAN and CycleGAN are first used to transfer the 650 original flower paintings in the testing set of the CFP dataset to natural flower images. 
Then, this fine-tuned flower classifier is used to directly classify 650 original flower paintings and the transferred flower paintings that are generated by the DSTN and other models. 

\textbf{Results}{\;}
The classification accuracy on the 650 original flower paintings and the transferred flower paintings is shown in Table \ref{tab:classification}. 
As it can be observed, before the domain transfer, the accuracy of directly classifying paintings is very low due to the domain bias. 
However, after the domain transfer, the classification accuracy of the transferred paintings produced by the DSTN, 
the DualGAN \cite{yi2017dualgan} and the CycleGAN \cite{zhu2017unpaired} show considerable improvements. 
The GANs \cite{goodfellow2014generative} perform badly due to the mode collapse. 
The DualGAN and the CycleGAN show a similar performance as they have similar structures. 
The transferred paintings produced by the proposed DSTN achieve the highest accuracy rate and each loss term employed contributes to this good performance. 
The experiment results show that the flower painting classification problem is successfully shifted to be a natural flower image classification problem. 
In addition, the proposed DSTN is proved effective in generating authentic transferred flower paintings. 

\begin{table}[ht]
\fontsize{7.5}{10}\selectfont 
\caption{\label{tab:classification}Mean fine-grained flower classification accuracy on flower paintings and transferred flower paintings generated by different methods.}% Higher accuracy means a better performance of domain transfer.}
\begin{center}
\begin{tabular}{l|lr}
\toprule
 \textbf{Classifier} & \textbf{Model} & \textbf{Accuracy [\%]} \\
 \midrule
%    \multirow{1}{*}{ResNet-152}  &Original paintings&31.08\\
%     & (Before Transfer) &     \\ 
 %  \midrule
    \multirow{7}{*}{ResNet-152}&GANs \cite{goodfellow2014generative}  &10.76\\
                                                   &DualGAN \cite{yi2017dualgan}   &86.75\\
                                                    &CycleGAN \cite{zhu2017unpaired}  &83.94\\
                                                    &CycleGAN+$L_{id}$ \cite{zhu2017unpaired}  &88.53\\
                                                     &DSTN (without $L_{cyc}$)  &19.68\\
                                                    &DSTN (without $L_{id}$)    &60.97\\
                                                     &DSTN (without $L_{con}$)    &88.97\\
                                                     &DSTN (Ours)     &\textbf{93.56} \\
\bottomrule
\end{tabular}
\end{center}
\end{table}

\subsection{Semantic Segmentation} \label{sec:segm}
\begin{figure*}[tb!]
\centering
\noindent\includegraphics[width=1.90\columnwidth]{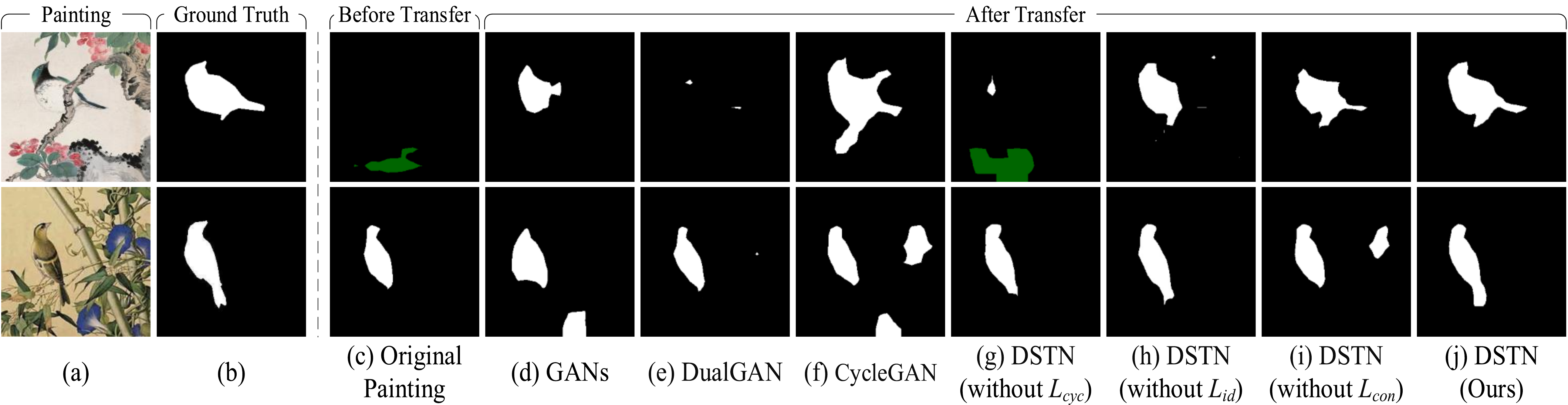}
\protect\caption{(a) Original painting, (b) Ground truth masks for original paintings, (c) Segmentation predictions of original paintings, (d$\sim$j) Segmentation predictions of the transferred paintings produced by different methods.}
\label{fig:seg}
\end{figure*}

In this section, semantic segmentation is conducted on the transferred bird paintings in order to verify the authenticity and quality of the transferred bird paintings. 
Specially, to better evaluate the results, segmentation masks of 600 bird paintings in the testing set of the CBP dataset are annotated by humans. 
Therefore, each painting in the testing set of the CBP dataset has a ground truth segmentation mask, as shown in Figure \ref{fig:seg} (b). 

\textbf{Setup and Dataset}{\;}The deeplab-V3 proposed in \cite{chen2017rethinking} is used in our segmentation model. 
The model adopts the ImageNet-pretrained \cite{deng2009imagenet} ResNet \cite{He2015Deep} to the semantic segmentation by applying atrous convolution to extract dense features. 
The deeplab-V3 is trained on the PASCAL VOC 2012 semantic segmentation benchmark \cite{everingham2015pascal} augmented by the extra annotations provided by \cite{hariharan2011semantic}. %resulting in 10,582 training images, 1,449 validation images and 1,456 testing images. 
There are 20 foreground object classes and 1 background class contained in the dataset, $e.g.,$ bird, cat, cow, dog, horse. 

\textbf{Experiments}{\;}
The trained models of the DSTN, GANs, DualGAN, CycleGAN are first used to transfer the 600 original bird paintings in the testing set of the CBP dataset to natural images. 
Then, the deeplab-V3 model is used to segment the birds. 
For a better comparison, the model is applied to both original bird paintings and transferred bird paintings produced by different methods. 
The performance is measured in terms of per-pixel accuracy and bird-class intersection-over-union (IOU). 

\textbf{Results}{\;}
The segmentation results of the \emph{$\{$bird painting $\rightarrow$ natural bird image$\}$} task are shown in Table \ref{tab:segm}. 
Moreover, Figure \ref{fig:seg} shows the visualization of the segmentation results of the transferred bird paintings. 
In Table \ref{tab:segm}, the pixel accuracy and IOU of the bird class show improvements after the domain transfer by the DSTN and the CycleGAN. 
The DSTN performs better than other state-of-art generative models in generating photo-realistic bird images. 
Additionally, it can be observed that the DSTN without $L_{id}$ shows the best performance. 
It is suspected that this may be because the $L_{con}$ gets a relatively higher weight when the $L_{id}$ is not used and the $L_{con}$ helps to maintain the original content, resulting in more detailed and clear images. 
From Figure \ref{fig:seg}, it can be seen that before the domain transfer, the bird areas in the paintings are mis-segmented by the deeplab-V3 trained on the natural image domain, 
$e.g.,$ the green part in column (c). The model deeplab-V3 shows a bad performance due to the domain bias. 
However, after domain transfer by the DSTN, the birds can be clearly segmented. 
The segmentation results also verify the authenticity and quality of the transferred bird paintings generated by the DSTN and the effectiveness of the proposed method. 
\begin{table}[htbp]
\centering
\fontsize{7.5}{10}\selectfont
\caption{\label{tab:segm}Segmentation accuracy for the \emph{$\{$bird painting $\rightarrow$ natural bird image$\}$} task. Higher is better. }
\begin{tabular}{lrr}
\toprule
\textbf{Model} & \textbf{Pixel Acc [\%]} & \textbf{IOU (bird) [\%]} \\
\cmidrule{1-3}
GANs \cite{goodfellow2014generative}                                                                                            & {80.40}    & {34.86}    \\
DualGAN \cite{yi2017dualgan}                                                                                       & {79.69}    & {38.72}     \\
CycleGAN \cite{zhu2017unpaired}                                                                                     & {80.27}    & {40.47}    \\
CycleGAN+$L_{id}$ \cite{zhu2017unpaired}                                                                     & {89.65}    & {49.97}    \\
\midrule
DSTN (without $L_{cyc}$)                                                            & {91.08}    & {54.10}      \\
DSTN (without $L_{con}$)                                                            & {91.34}    & {52.13}      \\
DSTN (without $L_{id}$)                                                               & {92.34}     & {55.43}    \\
DSTN(Ours)                                                                                    & {92.06}    & {54.36}     \\

\bottomrule
\end{tabular}
\end{table}

\subsection{Human Perceptual Study}\label{sec:human_study}
We conducted "real vs fake" human perceptual studies on \emph{$\{$flower painting $\rightarrow$ natural flower image$\}$}, \emph{$\{$bird painting $\rightarrow$ natural bird image$\}$} and \emph{$\{$landscape painting $\rightarrow$ natural landscape image$\}$} tasks. 
The perceptual protocol follows \cite{isola2016image}.  
100 participants consisted of undergraduate and graduate students from 7 different majors and they were given a series of images, 
which included real natural images and fake images generated either by our network or by other generative models. 
All images were shown at $256\times 256$ resolution and each image appeared for 1 second. 
Then, the image disappeared and participants were asked to select the images they believed were real and given unlimited time to respond. 
Each participant did 150 trials which contained 50 real images and 100 fake images produced by two different methods. 
The test was run under identical conditions for each method and 50 testing images of each method were randomly chosen from its transferred paintings. 

\begin{table}[htbp]
\centering
\fontsize{7.5}{10}\selectfont
\caption{\label{tab:human_study}Results of Human Perceptual Study. "real vs fake" test on the transferred bird, flower and landscape paintings generated by the different methods. Shown below is the percentage [\%] of trials in which participants were deceived by fake images. A higher value means more convincing fake images.}
\begin{tabular}{lrrr}
\toprule
\textbf{Model} & \textbf{Flower} & \textbf{Bird} & \textbf{Landscape}\\
\cmidrule{1-4}

 Real Natual Images     & 98.91                  & 99.62            & 97.40               \\
\midrule
GANs \cite{goodfellow2014generative}                                     & 2.33                 & 2.05                 & 21.96                \\
DualGAN \cite{yi2017dualgan}                                                    & 13.31                & 10.24                & 9.76                 \\ 
CycleGAN \cite{zhu2017unpaired}                                            & 12.44                & 12.15                & 10.20                \\
CycleGAN + $L_{id}$ \cite{zhu2017unpaired}                           & 30.44                & 20.17                & 18.25               \\
\midrule
DSTN (without $L_{cyc}$)                                                            & 10.15               &19.3             & 19.5                \\ 
DSTN (without $L_{con}$)                                                            & 32.71              & 27.23               & 24.06                  \\
DSTN (without $L_{id}$)                                                               & 21.34                & 19.76               & 23.20            \\
DSTN(Ours)                                                                                    & \textbf{43.66}              &  \textbf{28.66}               &  \textbf{44.78}                \\ 
\bottomrule
\end{tabular}
\end{table}

Table \ref{tab:human_study} shows the results of the human perceptual test. 
As it can be observed, the images generated by the DSTN fooled most of the participants, especially the transferred flower and landscape paintings. 
In contrast, the images generated by other methods showed a worse performance on this test. 
These results indicate that the transferred paintings produced by the DSTN are plausible and photo-realistic, showing a promising applicability of the DSTN. 

\section{Conclusion}
In this work, 
we propose a novel solution for ancient painting processing problems which is to use the domain transfer technique to convert ancient paintings to natural images, 
so that the models trained on natural images can be directly applied to transferred paintings. 
We first collect three ancient painting datasets consisting of flower, bird and landscape paintings respectively. 
Then, the proposed DSTN is used to transfer these paintings into photo-realistic natural images. 
Both qualitative and quantitative experiment results show the authenticity of the transferred paintings and the superiority of the proposed method. 
The proposed DSTN is not limited to the domain transfer between ancient paintings and natural images but could as well be applied to other domain transfer tasks where paired data is not available. 
The work provides a new way for art researchers to extract more information from paintings and promote the development of art research.

{\small
\bibliographystyle{ieee}
\bibliography{wacv2019}
}

\end{document}